\documentclass[10pt,twocolumn,letterpaper]{article}

\usepackage[pagenumbers]{cvpr} % To force page numbers, e.g. for an arXiv version
\usepackage[accsupp]{axessibility}
\usepackage{amsmath}    
\usepackage{amsfonts} 
\usepackage{multirow}
\usepackage{color}
\usepackage{graphicx}
\usepackage{makecell}
\usepackage{pifont}

\definecolor{cvprblue}{rgb}{0.21,0.49,0.74}
\usepackage[pagebackref,breaklinks,colorlinks,citecolor=cvprblue]{hyperref}
\def\paperID{30} % *** Enter the Paper ID here
\def\confName{CVPR}
\def\confYear{2024}

\title{Self-Supervised Class-Agnostic Motion Prediction with Spatial and Temporal Consistency Regularizations}

\author{
    Kewei Wang\textsuperscript{\rm 1,2},
    Yizheng Wu\textsuperscript{\rm 1,2},
    Jun CEN\textsuperscript{\rm 1},
    Zhiyu Pan\textsuperscript{\rm 2},
    Xingyi Li\textsuperscript{\rm 1,2},
    Zhe Wang\textsuperscript{\rm 3},
    Zhiguo Cao\textsuperscript{\rm 2},
    Guosheng Lin\textsuperscript{\rm 1}\thanks{Corresponding author.\\{gslin@ntu.edu.sg, \{wangkewei, zgcao\}@hust.edu.cn,}}\\
\textsuperscript{\rm 1} S-Lab, Nanyang Technological University\\
\textsuperscript{\rm 2} School of AIA, Huazhong University of Science and Technology\quad
\textsuperscript{\rm 3} SenseTime Research\\
}

\begin{document}

\maketitle
\begin{abstract}
The perception of motion behavior in a dynamic environment holds significant importance for autonomous driving systems, wherein class-agnostic motion prediction methods directly predict the motion of the entire point cloud. While most existing methods rely on fully-supervised learning, the manual labeling of point cloud data is laborious and time-consuming. Therefore, several annotation-efficient methods have been proposed to address this challenge. Although effective, these methods rely on weak annotations or additional multi-modal data like images, and the potential benefits inherent in the point cloud sequence are still underexplored.
To this end, we explore the feasibility of self-supervised motion prediction with only unlabeled LiDAR point clouds. Initially, we employ an optimal transport solver to establish coarse correspondences between current and future point clouds as the coarse pseudo motion labels. Training models directly using such coarse labels leads to noticeable spatial and temporal prediction inconsistencies. To mitigate these issues, we introduce three simple spatial and temporal regularization losses, which facilitate the self-supervised training process effectively. Experimental results demonstrate the significant superiority of our approach over the state-of-the-art self-supervised methods.
Code will be available at \url{https://github.com/kwwcv/SelfMotion}.
\end{abstract}

\begin{figure}[t]
\centering
\begin{subfigure}{.32\linewidth}
  \centering
  \includegraphics[width=1.0\linewidth]{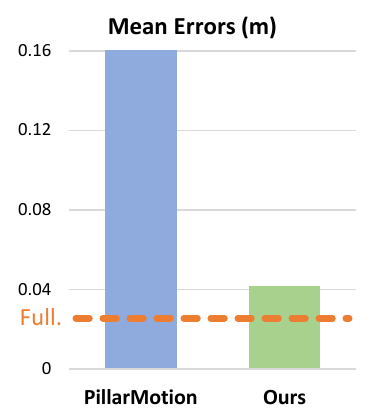}
  \caption{Static}
\end{subfigure}
\hfill
\begin{subfigure}{.32\linewidth}
  \centering
  \includegraphics[width=1.0\linewidth]{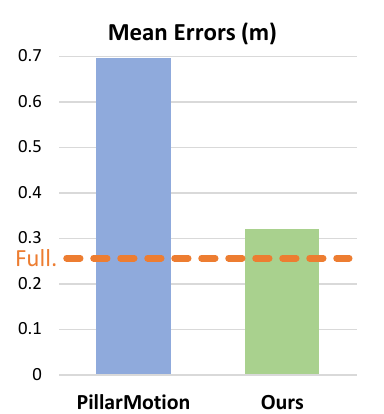}
  \caption{Slow}
\end{subfigure}
\hfill
\begin{subfigure}{.32\linewidth}
  \centering
  \includegraphics[width=1.0\linewidth]{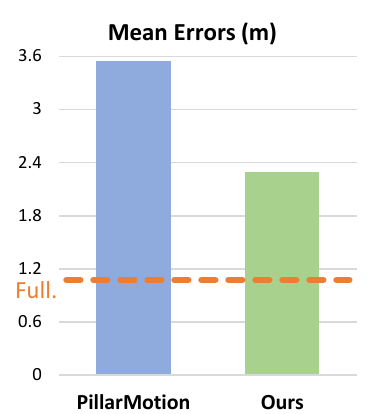}
  \caption{Fast}
\end{subfigure}

\caption{Performance comparison over static, slow, and fast speed levels between self-supervised PillarMotion~\cite{Luo2021SelfSupervisedPM} and our approach on the nuScenes dataset. The dashed line represents the performance of fully-supervised MotionNet~\cite{Wu2020MotionNetJP}. Our proposed self-supervised approach outperforms the PillarMotion which uses additional image data by a large margin and substantially narrows the performance gap with fully-supervised results.}
\label{fig:fig1}
 \vspace{-10pt}
\end{figure}

\section{Introduction}
Motion perception provides key information for autonomous driving. Classical approaches formulate the motion perception task as object-level trajectory prediction~\cite{chang2019argoverse, djuric2020uncertainty,fang2020tpnet, liang2020pnpnet, Zhao2020TNTTT}. However, these approaches may fail if agents are not successfully detected by the detectors, especially when encountering unseen categories in open-set scenarios~\cite{Wu2020MotionNetJP}. 
To compensate for trajectory prediction and provide backup, class-agnostic motion prediction was studied~\cite{LSTMED,Wu2020MotionNetJP,Wang2022BESTISI, weakmotion,semicm}. Currently, these approaches predict the motions of the entire environment from point clouds directly instead of for each agent.

While fully-supervised class-agnostic motion prediction methods have shown success, they typically require a vast amount of annotated point cloud data, which can be both costly and time-consuming to acquire. To overcome this limitation, WeakMotionNet~\cite{weakmotion} proposes a weakly supervised approach using only a limited number of foreground-background semantic annotations. WeakMotionNet also introduces a consistency-aware chamfer loss that uses the difference between correspondences from different frames as loss weights. Although effective, it requires establishing correspondences multiple times and is still based on matching results between individual frames, rather than taking advantage of the temporal information within the input sequences. 
Without using any annotations, PillarMotion~\cite{Luo2021SelfSupervisedPM} introduces a self-supervised training approach that incorporates regularization of 2D and 3D flows derived from paired image and point cloud data.
However, PillarMotion relies on multi-modality pairs to regularize motion learning, making the training process less flexible, and a considerable performance gap still exists between self-supervised and fully-supervised methods. 
Therefore, we explore the spatial and temporal information from the point cloud sequences themselves to train a better class-agnostic motion prediction model with only unlabeled LiDAR data in this paper.

Our approach primarily involves pseudo label generation and spatial and temporal regularizations. To generate pseudo motion labels, we first solve a matching problem to find correspondence between current and future point clouds. However, the correctness of the matching correspondence cannot be guaranteed due to the occlusion, background artifacts, noise, etc.  
Self-supervised learning from these coarse matching correspondences (pseudo labels) leads to two major failure cases: (1) predicted motions from the same rigid object are spatially diverse; (2) the background cells that should be static are predicted with large motions. 
To alleviate these issues, we introduce three simple but effective spatial and temporal regularizations from the point cloud motion prediction itself to facilitate self-supervised learning. 
Specifically, cluster consistency regularization, backward consistency regularization, and forward consistency regularization are proposed to encourage spatial consistency, penalize incorrect predictions (\eg, large motion for static cells), and smooth the prediction learned from pseudo labels of different time horizons, respectively. 

Experiments demonstrate that our proposed spatial and temporal consistency regularization terms work effectively together, significantly boosting the performance of self-supervised motion prediction. The proposed approach turns out with excellent performance, as illustrated in Fig.~\ref{fig:fig1}.
On the nuScenes dataset, our approach outperforms previous methods by a substantial margin.
Our contributions can be summarized as follows:
\begin{itemize}

\item  We propose a novel approach for self-supervised class-agnostic motion prediction that incorporates an optimal transport solver and three simple yet effective spatial and temporal regularization losses. 

\item Our approach is model-independent, and does not require any additional models or multi-modal data.

\item  Our method surpasses the state-of-the-art (SOTA) self-supervised methods by a large margin and significantly reduces the performance gap with the fully-supervised methods.

\end{itemize}

\begin{figure*}[t]
	\begin{center}
		\includegraphics[width=1.0\linewidth]{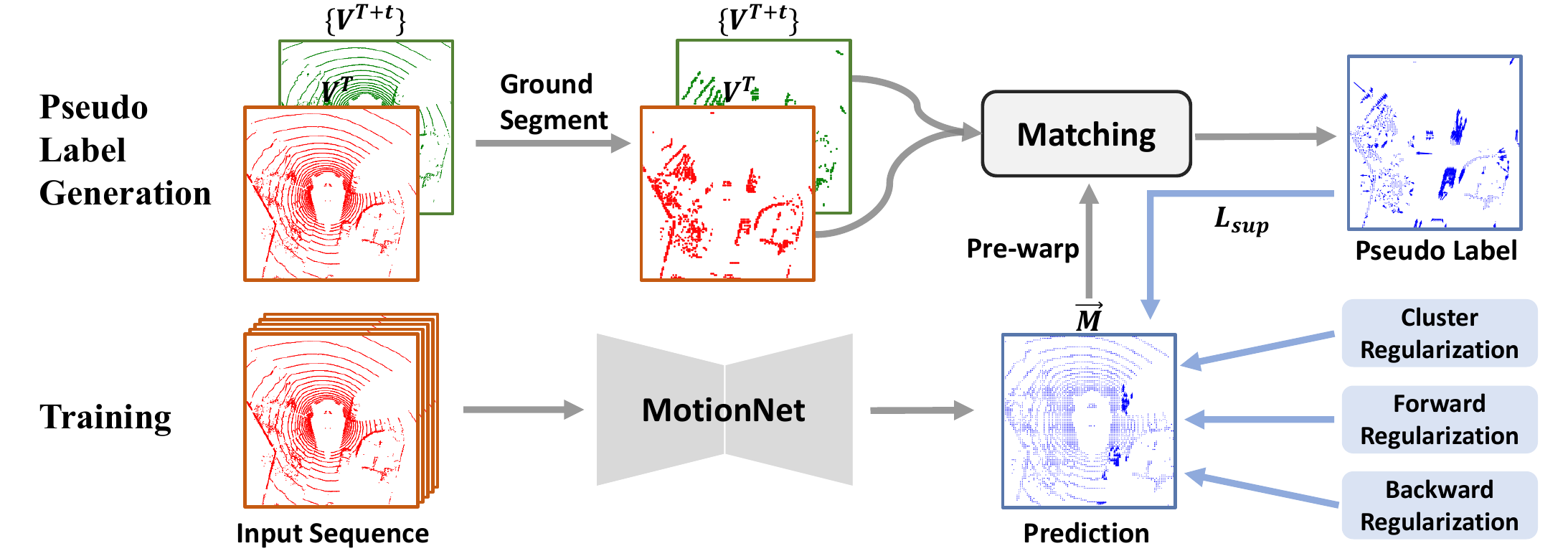}
		\caption{Overview of the proposed approach. Without ground truth labels, we first generate pseudo labels by matching. We then introduce cluster, forward, and backward regularization losses to facilitate self-supervised motion learning.}
		\label{fig:flowchart}
	\end{center}
 \vspace{-20pt}
\end{figure*}

\section{Related Works}
\noindent\textbf{Class-agnostic motion prediction.} Classical motion prediction approaches aim to predict the future trajectories of agents based on past observations. The entire system typically involves object detection~\cite{Zhou2017VoxelNetEL, Luo2018FastAF, lang2019pointpillars}, tracking, and forecasting~\cite{chang2019argoverse, djuric2020uncertainty,fang2020tpnet, liang2020pnpnet, Zhao2020TNTTT,ye2023bootstrap}. However, the trajectory prediction relies on the object detector which may fail to detect objects with unknown classes in open-set scenarios~\cite{Wu2020MotionNetJP}. 
To enhance system redundancy, especially in cases where objects may be not successfully detected and potentially lead to a failure in object-level trajectory forecasting~\cite{Wu2020MotionNetJP}, class-agnostic motion prediction methods predict the motion of entire scenes directly instead of individually for each agent. These methods represent the environment state based on BEV maps discretized from point clouds and aim to predict the 2D displacement vector for each BEV cell along the ground plane. 
MotionNet~\cite{Wu2020MotionNetJP} is one of the pioneers in this task, proposing to perform joint perception and motion prediction based on a spatial-temporal pyramid network. LSTM-ED~\cite{LSTMED} introduce convolutional LSTM~\cite{Shi2015ConvolutionalLN} to aggregate temporal context. 
% PillarFlow~\cite{lee2020pillarflow} adapts an optical flow network from~\cite{sun2018pwc} to learn motion flow between two consecutive BEV maps.
BE-STI~\cite{Wang2022BESTISI} leverages semantic cues by training an additional semantic decoder to guide motion prediction.

To alleviate the reliance on manual annotations, WeakMotionNet~\cite{weakmotion} proposes to train the MotionNet in a weakly-supervised manner with limited foreground and background semantic annotations.
PillarMotion~\cite{Luo2021SelfSupervisedPM} proposes to train the model in a self-supervised manner without any annotations by cross-sensor motion regularization. 
Although effective, a substantial performance gap persists between the self-supervised PillarMotion and the fully-supervised model even using additional image data. 
Therefore, an open question remains about whether we can more efficiently train a self-supervised motion prediction model with higher performance.

\noindent\textbf{Scene flow estimation.}
Scene flow estimation~\cite{FlowNet3D,Gu2019HPLFlowNetHP, pointpwc,behl2019pointflownet, wang2021hierarchical} from point clouds is an alternative way to understand the dense class-agnostic 3D motion field. 
In self-supervised scene flow estimation, several methods rely on establishing point correspondences between source and target point clouds through point matching. They treat the 3D coordinate difference between each matching pair as a pseudo scene flow label~\cite{puy2020flot, Li2021SelfPointFlowSS, mittal2020just, pointpwc, kittenplon2021flowstep3d, Li2022RigidFlowSS}. 
Many methods also utilize the chamfer distance loss~\cite{chamfer} to drive self-supervised scene flow learning~\cite{kittenplon2021flowstep3d, pontes2020scene, pointpwc,li2021neural,dong2022exploiting}.

While similar to the class-agnostic motion prediction task, scene flow estimation majorly differs in the following ways:
(1) The objectiveness of the two tasks is different~\cite{Wu2020MotionNetJP,weakmotion}. Scene flow estimation methods focus on estimating the motion flow between two observed point clouds, while motion prediction methods aim to forecast future displacements based on past observations; (2) Estimating dense 3D flow is time-consuming, which is not feasible for self-driving systems~\cite{weakmotion}.
Furthermore, directly applying scene flow estimation on real LiDAR point clouds poses challenges due to the lack of one-to-one correspondences~\cite{Wang2022BESTISI, Luo2021SelfSupervisedPM}.

\section{Method}
In this section, we begin by introducing the problem formulation of class-agnostic motion prediction. Subsequently, we delve into a comprehensive presentation and discussion of each component comprising our approach.
The overview of our approach is illustrated in Fig.~\ref{fig:flowchart}.
\subsection{Problem Formulation}
Given a temporal sequence of LiDAR point cloud frames, past frames are first synchronized to the current coordinate system~\cite{Wu2020MotionNetJP}. We denote synchronized point cloud captured at time $t$ as $P^t$. Following ~\cite{Wu2020MotionNetJP}, we discretized $P^t$ into dense voxels $V^t\in \{0,1\}^{H\times W \times C}$, where 0 indicates the voxel is empty, 1 indicates the voxel is occupied by at least 1 point, and $H$, $W$, $C$ are the voxel numbers along $X$, $Y$, $Z$ axis respectively.
Then we represent the 3D voxel lattice as a 2D pseudo-image with the $C$ corresponding to the image channel and view $V^t$ as a virtual BEV map. Then the motion field $M \in \mathbb{R}^{H\times W\times 2}$ in the BEV map is defined as the 2D displacement of each grid cell to the future timestamp. Taking sequence $\overrightarrow{\mathcal{V}}=\{V^t\}_{t=1}^T$ as input, the task aims to predict the motion fields $\overrightarrow{\mathcal{M}}=\{M^{T\rightarrow T+t}\}_{t=1}^{T'}$, where $T$ and $T'$ are the number of past and future timestamps, respectively.

\subsection{Overview}
As depicted in Fig.~\ref{fig:flowchart}, our approach mainly involves pseudo label generation and self-supervised training. 
To generate pseudo labels, we initially solve the matching problem between two consecutive frames. Firstly, we remove the majority of ground plane points from point clouds by ground segmentation algorithm \cite{lee2022patchwork++} to reduce noise for better matching. 
Subsequently, we feed the raw BEV map sequence into MotionNet~\cite{Wu2020MotionNetJP} and train the model using the generated pseudo labels through supervised loss $\mathcal{L}_{sup}$. As the quality of pseudo labels is low at the initial stage of training, we introduce three spatial and temporal consistency regularization losses, named cluster consistency loss $\mathcal{L}_{c}$, forward consistency loss $\mathcal{L}_{f}$, and backward consistency loss $\mathcal{L}_{b}$, to facilitate the self-supervised learning.

\begin{figure*}[t]
\centering
\begin{subfigure}{.28\linewidth}
  \centering
  \includegraphics[width=1.0\linewidth]{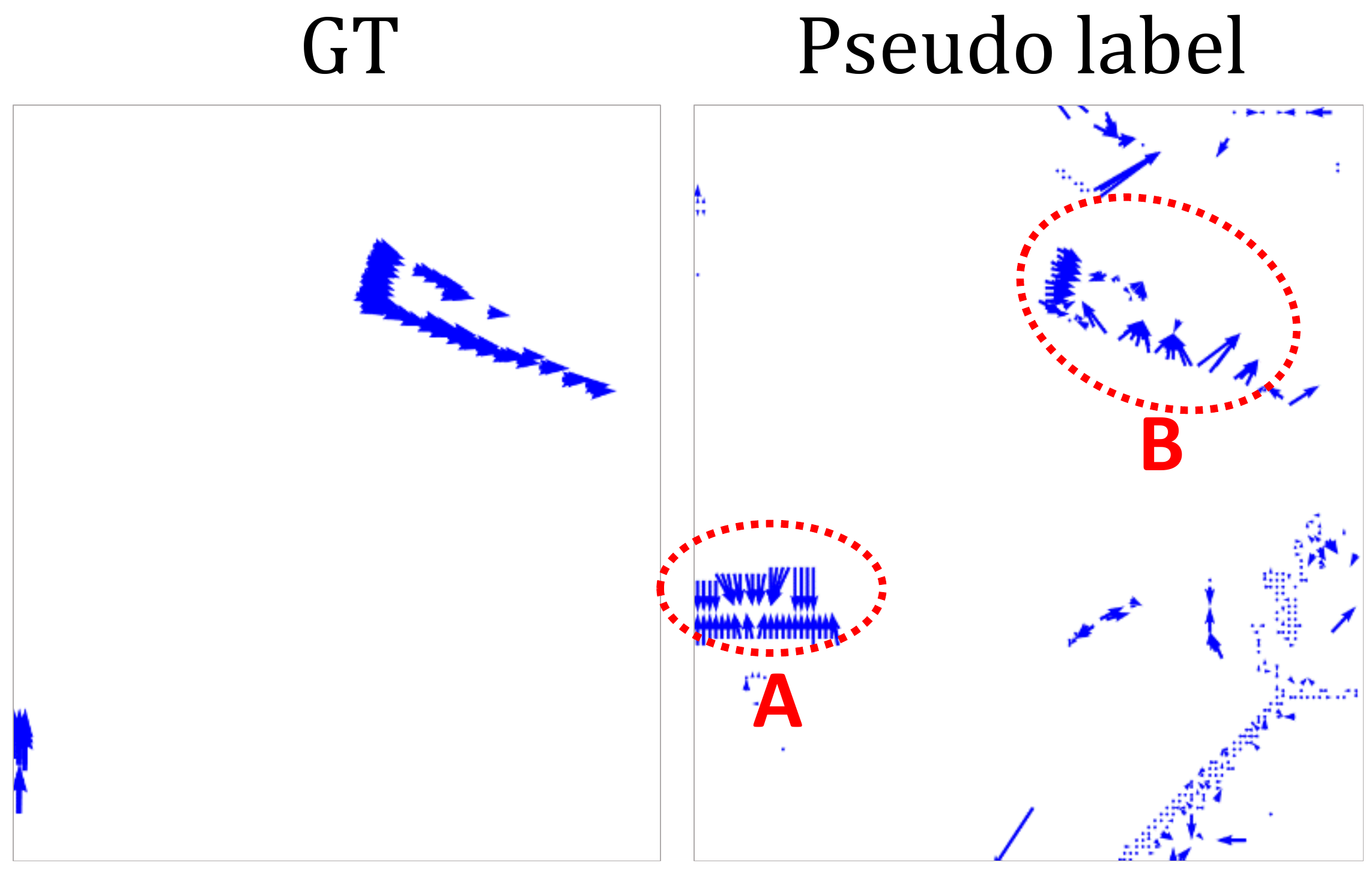}
  \caption{GT \& Pseudo label}
  \label{fig:methoda}
\end{subfigure}
\hfill
\begin{subfigure}{.14\linewidth}
  \centering
  \includegraphics[width=1.0\linewidth, height=1.28\linewidth]{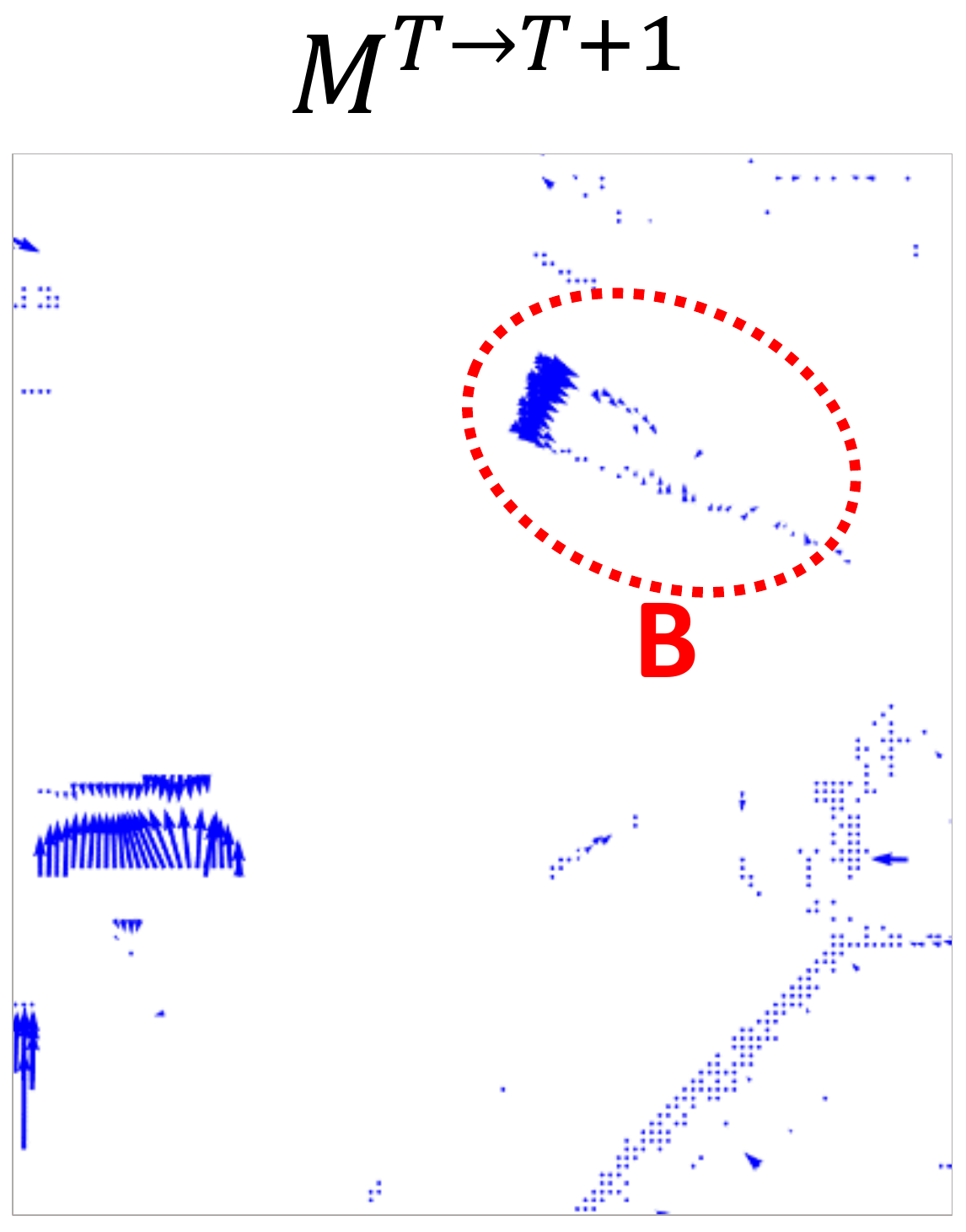}
  \caption{Spatial incons.}
  \label{fig:methodb}
\end{subfigure}
\hfill
\begin{subfigure}{.28\linewidth}
  \centering
  \includegraphics[width=1.0\linewidth, height=0.635\linewidth]{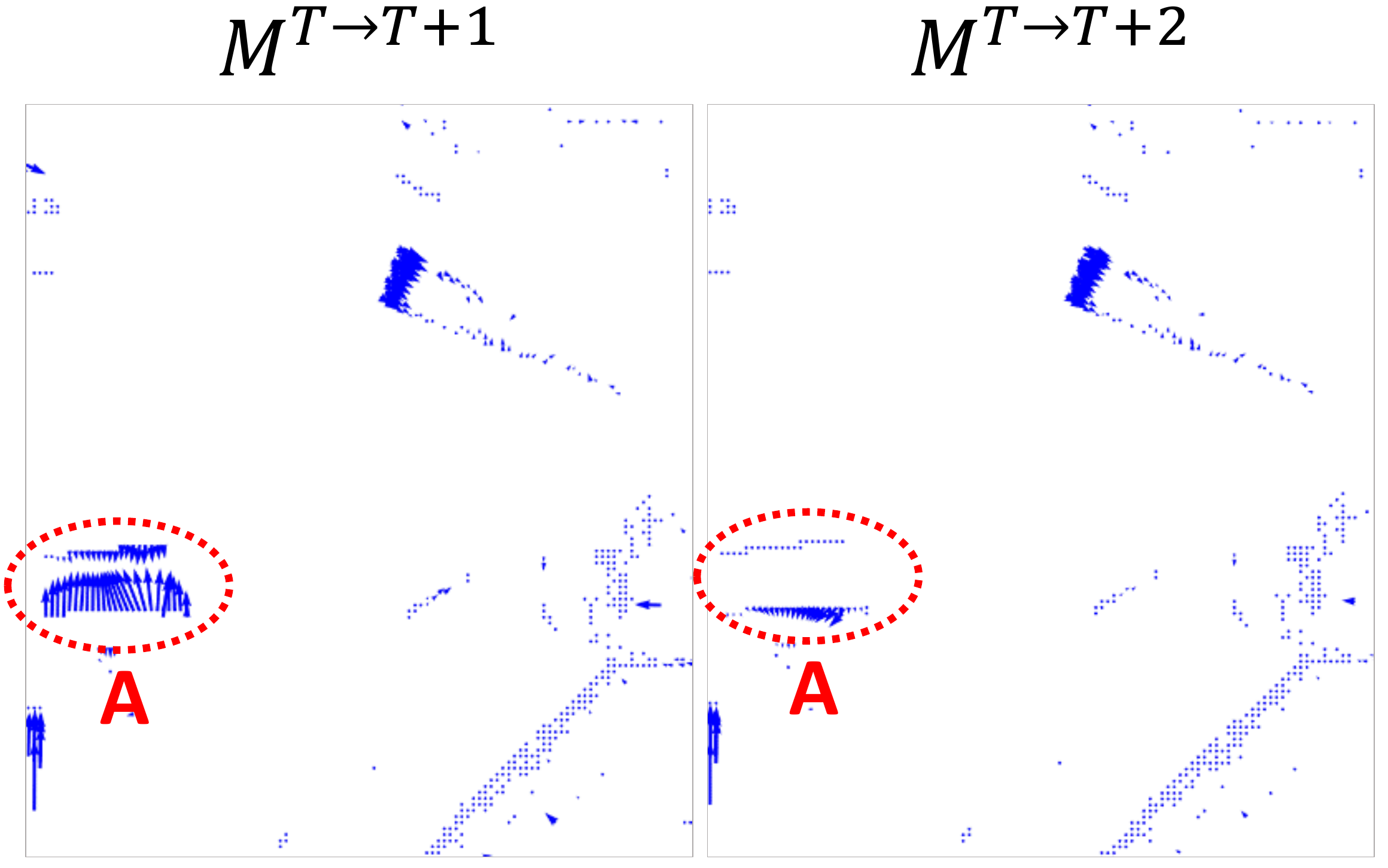}
  \caption{Forward temporal incons.}
  \label{fig:methodc}
\end{subfigure}
\hfill
\begin{subfigure}{.28\linewidth}
  \centering
  \includegraphics[width=1.0\linewidth]{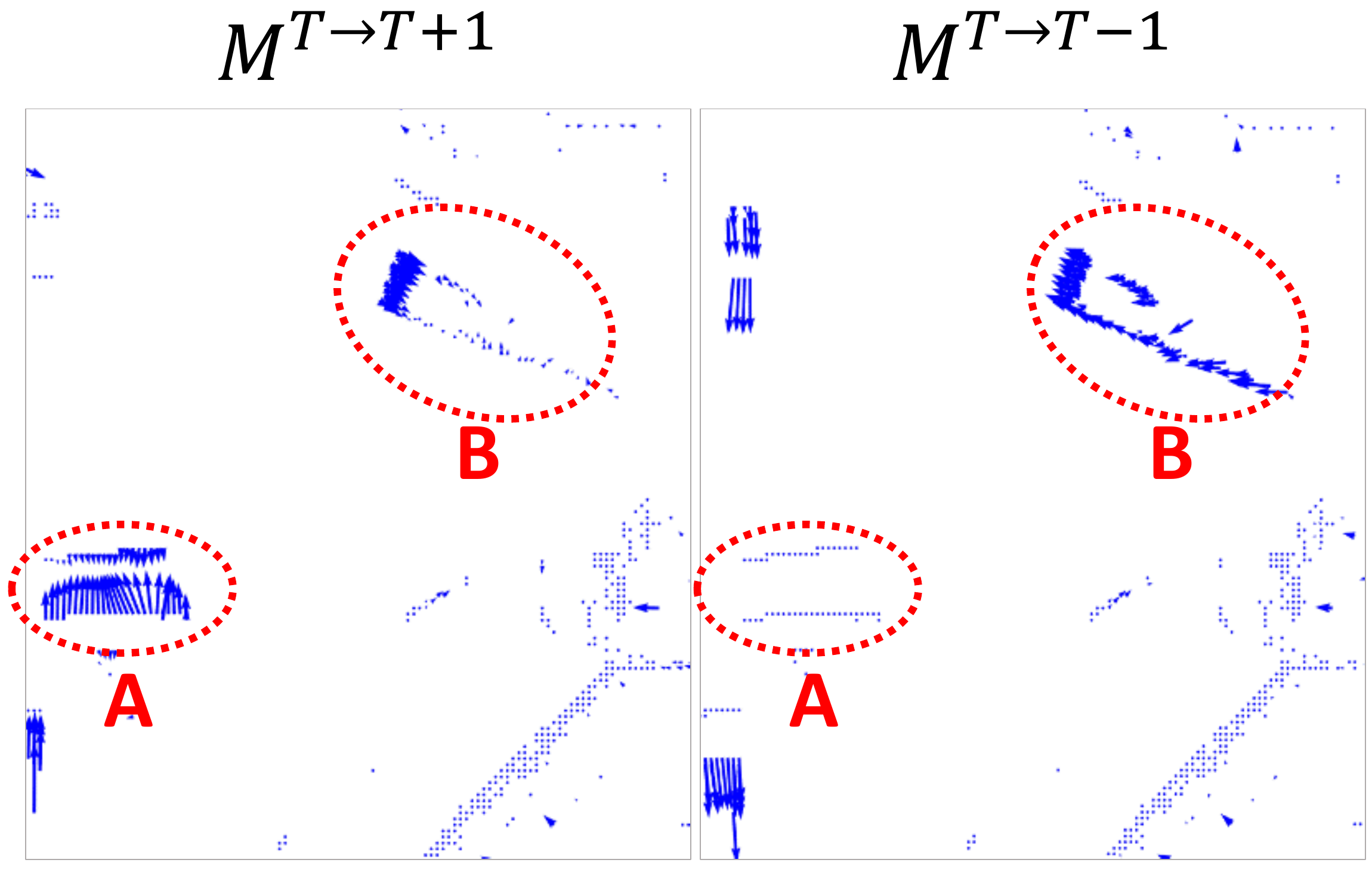}
  \caption{Backward temporal incons.}
  \label{fig:methodd}
\end{subfigure}

\caption{Prediction results with inconsistency. (a) Ground truth and pseudo labels. (b) Predictions from the same object are inconsistent. (c) Predictions for consecutive future timestamps are inconsistent. (d) Forward and backward predictions (\eg, $M^{T\rightarrow T+1}$ and $M^{T\rightarrow T-1}$) are inconsistent. Inconsistency regions are highlighted by red circles. The blue arrow denotes the future displacement (motion).}
\label{fig:method}
\end{figure*}

\subsection{Pseudo Label Generation}
\label{sec:matching}
Unlike fully-supervised learning, where ground truth labels are available, self-supervised methods learn from data itself without annotations. In this condition, one alternative approach is to generate pseudo labels to provide the necessary supervision for training.
% Unlike fully supervised learning, ground truth labels are not available in the self-supervised condition, and one alternative way is to generate pseudo labels for supervision.
Considering motion can be defined as the variation in coordinates between correspondent points in two consecutive frames, the process of generating pseudo labels can be viewed as a matching problem. We denote $B^T=\{b^T_i \in \mathbb{R}^2 \}^{N_T}_{i=1}$ as the 2D coordinates of BEV cells on the current virtual BEV map $V^T$, where $N_T$ is the number of non-empty cells. Then we perform matching between source $B^T$ and target $B^{T+t}=\{b^{T+t}_j \in \mathbb{R}^2 \}^{N_{T+t}}_{j=1}$. Ideally, if there exists one-to-one correspondence between $B^T$ and $B^{T+t}$, $B^T$ can be projected into $B^{T+t}$ and fully occupy $B^{T+t}$:
\begin{equation}
    B^T + M = \pi B^{T+t},
\label{eq:pseudo}
\end{equation}
where $\pi \in \{0,1\}^{N_T\times N_{T+t}}$ is a permutation matrix indicating correspondence between $B^T$ and $B^{T+t}$, and $M$ is the motion flow. The optimal soft permutation matrix $\pi^* \in \left[0,1\right]^{N_T\times N_{T+t}}$ can be computed by solving an optimal transport problem:
\begin{equation}
\begin{split}
    &\qquad\qquad\pi^* = \mathop{\arg\min}\limits_{\pi} \sum_{i,j}C_{ij}\pi_{ij}\\
    &s.t.\quad \pi \mathbf{1}_{N_{T+t}} = \frac{1}{N_T} \mathbf{1}_{N_T}, \pi^T \mathbf{1}_{N_T} = \frac{1}{N_{T+t}} \mathbf{1}_{N_{T+t}},
\end{split}
\end{equation}
where $\mathbf{1}_{N_T}\in \mathbb{R}^{N_T}$ and $\mathbf{1}_{N_{T+t}}\in \mathbb{R}^{N_{T+t}}$ are vectors with all elements are 1. $C$ is the transport cost matrix and $\pi^T$ is the transpose of $\pi$. This problem can be solved by the Sinkhon algorithm~\cite{cuturi2013sinkhorn}. We use a pre-warping operation to warp the source BEV cells $B^T$ by the predicted motion $M^{T\rightarrow T+t}$, and find the correspondence between the warped BEV cells $\hat{B}^T = B^T + M^{T\rightarrow T+t}$ and target $B^{T+t}$ to calculate $C$. In this way, with better predictions, the quality of pseudo labels keeps improving during training. The cost matrix $C$ is obtained by computing the pairwise distance between coordinates:
\begin{equation}
    C_{ij} = 1-\exp{(-\frac{||\hat{b}_i^T-b_j^{T+t}||^2}{\theta_c})},
\end{equation}
where $\theta_c$ is a temperature parameter.
According to Eq.\ref{eq:pseudo}, the pseudo motion labels can be obtained by:
\begin{equation}
    \hat{M}^{T \rightarrow T+t} = \pi^* B^{T+t} - B^T.
% \label{eq:pseudo}
\end{equation}
With the pseudo motion labels, the supervised training loss can be computed by:
% \begin{equation}
%     \mathcal{L}_{sup} = smooth_{L1} (\hat{M}^t, M^t)
% \end{equation}
\begin{equation}
    \mathcal{L}_{sup} = \sum_{t=1}^{T'}smooth_{L1} (M^{T \rightarrow T+t}, \hat{M}^{T \rightarrow T+t})
\end{equation}
During pseudo label generation, we first employ a training-free ground segmentation algorithm~\cite{lee2022patchwork++} to remove the majority of ground plane points to reduce the negative influence of ground points for matching and set the motion labels of these removed points as zeros. $\theta_c$ is set as 3 in the paper.

\subsection{Spatial-Temporal Consistency Regularizations}
\subsubsection{Motivations}
Due to factors such as occlusion, ground points, and sensors moving, establishing correct correspondences between source and target BEV maps is challenging. Despite removing ground points by the ground segmentation algorithm, noises without correspondences remain. The existence of these noises and cells without correspondences significantly impairs the solution to optimal transport problems. Consequently, the generated pseudo labels are inaccurate, leading to suboptimal outcomes. 

For instance, certain cells corresponding to static background regions may get apparent moving labels (\eg, circle A in Fig.~\ref{fig:method} (a)), while cells associated with moving objects might possess static or inaccurate moving labels (\eg, circle B in Fig.~\ref{fig:method} (a)).
Training the model only using such low-quality pseudo labels yields unsatisfactory results including the spatial and temporal inconsistent predictions, as shown in Fig.~\ref{fig:method} (b)(c)(d). Therefore, we introduce three spatial and temporal consistency regularization terms. These regularization terms aim to penalize inaccurate predictions and propagate accurate ones. 
And through the pre-warping operation, the quality of the pseudo labels can improve iteratively as the accuracy of predictions increases.

\subsubsection{Spatial cluster consistency regularization.}
In real-world traffic scenarios, the majority of objects, such as cars and trucks, exhibit a high degree of rigidity. It is intuitive to expect that cells belonging to the same rigid object should display consistent predicted motions.
Unfortunately, the generated pseudo labels for cells within the same rigid object are inconsistent (\eg, circle B in Fig.~\ref{fig:method} (a)). 
Consequently, directly learning from these pseudo labels may result in situations like half of the object being predicted with moving status but the other half being predicted with static status (\eg, circle B in Fig.~\ref{fig:method} (b)). Therefore, encouraging the consistency of predicted motion within the same object becomes an important factor in enhancing the performance of the self-supervised model.

Due to the unavailability of instance segmentation labels, one potential solution is to employ K-Nearest-Neighbors (KNN) consistency loss~\cite{pointpwc}. Although this loss helps encourage local smoothness, it does not consider instance-level consistency, especially in the case of large instances.
To this end, we utilize cluster consistency loss to perform the spatial consistency constraint. As cells belonging to the same object are often in close proximity to one another in the virtual BEV map, objects can be coarsely represented as compact clusters. Therefore, we first utilize a breadth-first clustering algorithm to cluster every single cell within a distance $d_c$ (The detailed clustering process is in the Supplementary). Then for each cluster, we encourage spatial consistency. 
As illustrated in Sec.~\ref{sec:matching}, we perform clustering on 2D BEV maps to reduce the computational cost. The cluster consistency loss is formulated as:
\begin{equation}
    \mathcal{L}_{c} = \frac{1}{|S|}\sum_{s_i\in S} \frac{1}{|s_i|^2}\sum_{b_i,b_j \in s_i}\left(||\overrightarrow{\mathcal{M}}\left(b_i\right)-\overrightarrow{\mathcal{M}}\left(b_j\right)||\right),
\end{equation}
where $S=\{s_i\}_{i=1}^{N_s}$, $s_i$ indicates a cluster, $N_s$ is the cluster numbers in the BEV map and $\overrightarrow{\mathcal{M}}\left(b\right)$ is the motion field at cell $b$. $d_c$ is set as 3 in the paper.  

\subsubsection{Temporal forward consistency regularization.}
From supervised loss $\mathcal{L}_{sup}$, the motions are learned individually from the pseudo labels for each timestamp. However, since the foreground and noise distributions of future BEV maps are different (\eg $B^{T+1}$ and $B^{T+2}$), we will get inconsistent pseudo motion labels between $\hat{M}^{T\rightarrow T+1}$ and $\hat{M}^{T\rightarrow T+2}$. 
As the model learns from these inconsistent pseudo labels, it will produce inconsistent predictions. For example, the predictions of static cells in ${M}^{T\rightarrow T+1}$ are wrong but correct in ${M}^{T\rightarrow T+2}$ ( circle A in Fig.~\ref{fig:method} (c)). As the object displacements in two consecutive short time windows are supposed to be consistent, we regularize forward predictions as:
\begin{equation}
    \mathcal{L}_{f}=\sum_{t=1}^{T'-1}smooth_{L1} (M^{T\rightarrow T+t}, \frac{t}{t+1} M^{T\rightarrow T+t+1}).
\end{equation}
With forward consistency loss $\mathcal{L}_f$, the predicted motions of adjacent timestamps act as mutual regularization, which is important for self-supervised motion learning as the pseudo labels are temporally inconsistent. For example, ${M}^{T\rightarrow T+2}$ can provide more accurate information for ${M}^{T\rightarrow T+1}$ in Fig.~\ref{fig:method} (c) case. 

\begin{table*}[t]
\centering
\caption{Performance comparison on nuScenes dataset. Full., Weak., and Self. refer to fully-supervised, weakly-supervised, and self-supervised training, respectively. Our approach surpasses state-of-the-arts by a large margin.}
\renewcommand\tabcolsep{10pt}
\renewcommand\arraystretch{1.2}
\resizebox{\linewidth}{!}{
\begin{tabular}{l|ccc|cccccc}
\hline
\multirow{2}*{Method} &\multirow{2}*{Sup.}&\multirow{2}*{Modality} &\multirow{2}*{Years} &\multicolumn{2}{c}{Static }&\multicolumn{2}{c}{Speed $\leq$ 5m/s (Slow)}&\multicolumn{2}{c}{Speed $\geq$ 5m/s (Fast)} \\
\cline{5-10}
&&&&Mean$\downarrow$ &Median$\downarrow$ &Mean$\downarrow$ &Median$\downarrow$ &Mean$\downarrow$ &Median$\downarrow$\\
\hline
Static &- &LiDAR &-&0 &0 &0.6111 &0.0971 &8.6517 &8.1412\\
FlowNet3D~\cite{FlowNet3D} &Full. &LiDAR &CVPR2018&0.0410 &0 &0.8183 &0.1782 &8.5261 &8.0230\\
HPLFlowNet~\cite{Gu2019HPLFlowNetHP} &Full. &LiDAR &CVPR2019&0.0041 &0.0002 &0.4458 &0.0960 &4.3206 &2.4881\\
% PointRCNN &Full. &LiDAR &0.0204 &0 &0.5514 &0.1627 &3.9888 &1.6252\\
LSTM-ED~\cite{LSTMED} &Full. &LiDAR &ICRA2019&0.0358 &0 &0.3551 &0.1044 &1.5885 &1.0003\\
MotionNet~\cite{Wu2020MotionNetJP} &Full. &LiDAR &CVPR2020&0.0256 &0 &0.2565 &0.0962 &1.0744 &0.7332\\
BE-STI~\cite{Wang2022BESTISI} &Full. &LiDAR &CVPR2022&0.0220 &0 &0.2115 &0.0929 &0.7511 &0.5413\\
WeakMotion~\cite{weakmotion} &Weak.  &LiDAR &CVPR2023&0.0558 &0 &0.4337 &0.1305 &1.7823 &1.0887\\
\hline
PointPWC~\cite{pointpwc} &Self. &LiDAR&ECCV2020& 0.5264 & 0 &0.9423 &0.3225 &3.9530 &2.5149\\
RigidFlow~\cite{Li2022RigidFlowSS} &Self. &LiDAR&CVPR2022& 0.6361 & 0 &1.1836 &0.5666 &3.9439 &2.5781\\
PillarMotion~\cite{Luo2021SelfSupervisedPM} &Self. &LiDAR+Image&CVPR2021& 0.1620 & 0.0010 &0.6972 &0.1758 &3.5504 &2.0844\\
% WeakMotionNet~\cite{weakmotion} &Self. &LiDAR&CVPR2023& 0.3137 & 0 &0.6281 &0.1632 &2.9726 &1.7997\\
WeakMotion~\cite{weakmotion} &Self. &LiDAR&CVPR2023& 0.1919 & 0 &0.4510 &0.1150 &2.9715 &1.8822\\
Ours &Self. &LiDAR &-&\textbf{0.0419} &\textbf{0} &\textbf{0.3213} &\textbf{0.1061} &\textbf{2.2943} &\textbf{1.0508}\\
\hline
\end{tabular}
}
\label{tab:main_results}
\end{table*}

\subsubsection{Temporal backward consistency regularization.}
\label{subsec:back}
When generating pseudo labels, it is inevitable to assign noise and some ground source cells to the wrong target cells due to the absence of corresponding cells in the target point clouds. 
Wrong pseudo labels consequently lead the trained model to predict incorrect motion, especially wrongly predicting large displacements for background cells (\eg circle A in left Fig.~\ref{fig:method} (d)). This creates a contradiction between the model's expected behavior (which is to predict static motion for background cells based on the past frames) and the actual predictions learned directly from incorrect pseudo labels. Inspired by the contradiction, we propose backward consistency loss to penalize incorrect predictions. 

Specifically, besides forward sequence $\overrightarrow{\mathcal{V}}=\{V^1,V^2,...,V^T\}$, we also feed additional backward sequence $\overleftarrow{\mathcal{V}}=\{V^{2T-1},V^{2T-2},...,V^T\}$ to the network to predict the backward motion $\overleftarrow{\mathcal{M}}$. Compared to the forward sequence, the backward sequence has entirely different past frames. During training, without being directly constrained by pseudo labels like the forward motion, the backward motion reflects the model's general ability to perceive motion cues hidden in the temporal sequences. Considering that the forward and backward speeds of an object are opposites at the same moment, we formulate the backward regularization term as:
\begin{equation}
    \mathcal{L}_{b}=\sum_{t=1}^{T'}\exp(-\frac{t}{\theta_b})smooth_{L1} (M^{T\rightarrow T+t}, -M^{T\rightarrow T-t}),
    \label{label:back}
\end{equation}
where $\theta_b$ is the temperature parameter. We use the $\exp(-\frac{t}{\theta_b})$ term to dynamically adjust the loss weights, taking into account that as the time horizon increases, accurately predicting motion becomes more challenging, and the forward-backward divergence becomes less reliable.
This inconsistency becomes noticeable in cells with incorrect predictions (Fig.~\ref{fig:bf_div}), which demonstrates that backward regularization can effectively reflect the accuracy of the predictions. Thus, the backward regularization $\mathcal{L}_{b}$ generally penalizes incorrect predictions while having less impact on correct ones. $\theta_b$ is set as 10 in the paper. In Sec.~\ref{sec:exp}, we will show that the backward regularization effectively benefits the self-supervised motion learning process.

\noindent\textbf{Overall loss functions.} 
To summarize, the overall training objective is:
\begin{equation}
    \mathcal{L}_{total}=\mathcal{L}_{sup} + \alpha \mathcal{L}_{c} + \beta \mathcal{L}_{f} + \gamma \mathcal{L}_{b},
\end{equation}
where $\alpha$, $\beta$, and $\gamma$ are hyper-parameters, balancing the contribution of the different loss terms.
It is worth noting that all the regularization terms are only used for training.

\section{Experiments}
\label{sec:exp}
In this section, we begin by comparing our methods with state-of-the-art (SOTA) motion prediction methods. Subsequently, we conduct ablation studies to analyze the effectiveness of each individual component.

\noindent\textbf{Dataset.} We mainly evaluate the proposed approach on the large-scale self-driving dataset: nuScenes~\cite{Caesar2019nuScenesAM}, which contains 850 scenes with annotations. For fair comparisons, we follow previous works~\cite{Wu2020MotionNetJP, Luo2021SelfSupervisedPM, Wang2022BESTISI} to use 500 scenes for training, 100 scenes for validation, and 250 scenes for testing.

\noindent\textbf{Implementation details.} We follow the same pre-process setting in~\cite{Wu2020MotionNetJP}, where the point clouds are cropped in the range of [-32m, 32m]$\times$ [-32m, 32m]$\times$ [-3m, 2m] and the voxel size is set to 0.25m $\times $0.25m $\times$ 0.4m along XYZ axis. We take the current sweep and the past four sweeps as input. We use MotionNet as the baseline model for its effectiveness and flexibility following~\cite{weakmotion, Luo2021SelfSupervisedPM}. We train the model for 20 epochs and set the learning rate to be 0.002. Adam is used as the optimizer. We implement our model in Pytorch~\cite{paszke2019pytorch} with a single RTX 3090 GPU. Hyperparameters $\alpha$, $\beta$, and $\gamma$ are set as 0.05, 0.1, and 1 respectively. Following previous works, we take the sequence with 5 frames ($T=5$) as input and show the results of the next 1s ($T'=5$) in the following tables. The time interval between each frame is 0.2s.

\noindent\textbf{Evaluation metrics.} Following previous works~\cite{Wu2020MotionNetJP,Luo2021SelfSupervisedPM, Wang2022BESTISI}, we evaluate the mean and median errors at different speed levels by dividing the grid cells into 3 groups according to the ground truth speeds: static, slow ($\leq$ 5m/s) and fast ($\geq$ 5m/s). Errors are measured by $L_2$ distances between the predicted displacements and the ground truth displacements. 

\subsection{Comparison with State-of-the-Art Methods}
In Table~\ref{tab:main_results}, we show the results of the following methods: (1) A static model that assumes a static environment. (2) Scene flow estimation methods, including FlowNet3D~\cite{FlowNet3D}, HPLFlowNet~\cite{Gu2019HPLFlowNetHP}, PointPWC~\cite{pointpwc}, and RigidFlow~\cite{Li2022RigidFlowSS}. Following~\cite{Wu2020MotionNetJP}, we employ scene flow estimation methods by assuming linear dynamics. First, we predict the motion by estimating the flow from the current time to the past 0.4s. Then, we project the predicted flow onto the BEV map and to the future 1.0s for comparison. (3) Class-agnostic motion prediction methods, including LSTM-ED~\cite{LSTMED}, MotionNet~\cite{Wu2020MotionNetJP}, PillarMotion~\cite{Luo2021SelfSupervisedPM}, BE-STI~\cite{Wang2022BESTISI}, and WeakMotionNet~\cite{weakmotion}.

Compared to fully-supervised models, our approach outperforms FlowNet3D and HPLFowNet at both slow and fast speed levels, and it surpasses the weakly-supervised WeakMotionNet (with 1\% foreground and background masks~\cite{weakmotion}) at static and slow speed levels. In comparison with self-supervised methods, our approach achieves superior performance to  PillarMotion, with significant improvements of 74.1\%(-0.1201 m), 53.9\% (-0.3759 m), and 35.4\%(-1.2561 m) at static, slow, and fast speed levels, respectively. To further show the superiority of the proposed approach, we implement recent WeakMotionNet in a self-supervised manner by replacing its pre-segmentation network with the same ground segmentation algorithm~\cite{lee2022patchwork++} used in our approach. While self-supervised WeakMotionNet outperforms PillarMotion at slow and fast speed levels, our approach still surpasses it by a large margin.

\subsection{Ablation Studies}
\label{sec:ablation}
In this subsection, we evaluate the effectiveness of each component of our approach. As shown in Table~\ref{tab:ablation}, the base model trained only with pseudo labels does not perform well at all speed levels (Row 1).

\noindent\textbf{Effectiveness of cluster consistency regularization.}
When cluster consistency loss is not applied, the model exhibits noticeably poorer performance at the fast speed level (Compare Row 1 and Row 6). This disparity in performance can be attributed to the way the cost matrix for the optimal transport problem is computed based on the distance between cells. Consequently, the solver tends to find correspondences between cells that are in close proximity to each other. While this proximity-based approach has its merits, it presents challenges in generating long displacement pseudo labels that the model can learn from. As a result, during the early stages of training, there is a scarcity of correct pseudo labels for fast motion. By imposing constraints on cluster consistency, the model is encouraged to predict faster motion, even if it corresponds to a short displacement pseudo label. This is possible due to the presence of fast displacement predictions within the same cluster. With larger displacement and pre-warping, the optimal transport solver is more likely to find correct correspondence between two points at long distances. In Table~\ref{tab:cluster}, we present the results of KNN consistency loss and cluster consistency loss, both of which function as spatial smooth losses. Compared to $\mathcal{L}_{KNN}$, $\mathcal{L}_{c}$ bring more benefits comprehensively. 

\begin{figure}[t]
	\begin{center}
		\includegraphics[width=0.9\linewidth]{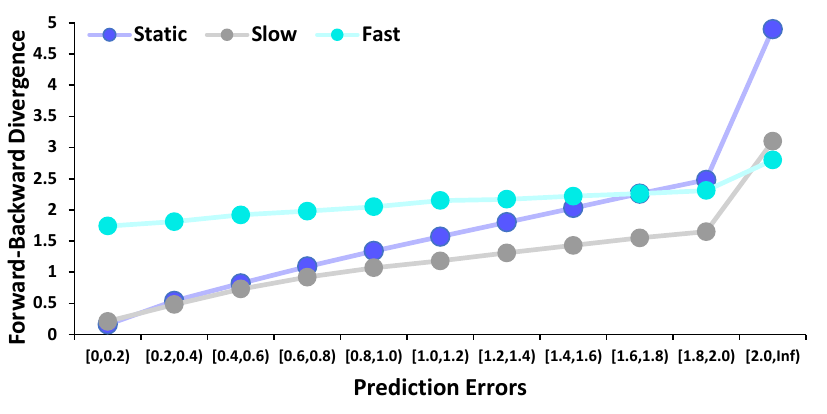}
		\caption{Forward-backward divergence of the training set. We analyze the prediction errors of all samples in the training set and their corresponding forward-backward divergences. The Forward-backward divergence is positively correlated with the prediction error.}
		\label{fig:bf_div}
	\end{center}
\end{figure}

\noindent\textbf{Effectiveness of temporal consistency regularization.}
Both forward and backward regularization terms boost the performances across all speed levels (Compare Row 5 and Row 7 with Row 8 respectively).
To further illustrate the impact of backward regularization, we separately feed forward and backward sequences of all training samples into the model only trained with $\mathcal{L}_{sup}$ to demonstrate prediction divergence ($|\overrightarrow{\mathcal{M}} -\overleftarrow{\mathcal{M}}|$).
As depicted in Fig.~\ref{fig:bf_div}, there is a positive correlation between the forward-backward prediction divergence and prediction accuracy. The divergence is large when the prediction is inaccurate and vice versa, which indicates that backward regularization will have a stronger impact on inaccurate predictions but a weaker influence on accurate ones. 
By using the forward-backward divergence as a loss term, the model is penalized for making inaccurate predictions. 
Results for various $\theta_b$ are shown in Table~\ref{tab:thetab}, which indicates that the weight term can further improve the performance.

\begin{table}[t]
\centering
\caption{Ablation study for different combinations of proposed regularization terms. Bold fonts and underlines indicate the best and the second-best performance, respectively.}
\renewcommand\tabcolsep{12pt}
\renewcommand\arraystretch{1.5}
\resizebox{\linewidth}{!}{
\begin{tabular}{p{1pt}p{0.5pt}p{0.5pt}p{0.5pt}p{0.5pt}|ccc}
\hline
\multirow{2}*{\#Row}&\multirow{2}*{$\mathcal{L}_{sup}$}&\multirow{2}*{$\mathcal{L}_{c}$} &\multirow{2}*{$\mathcal{L}_{f}$} &\multirow{2}*{$\mathcal{L}_{b}$}&Static &Slow &Fast \\
\cline{6-8}
&&&&&\multicolumn{3}{c}{Mean Error $\downarrow$}\\
\hline
1&\checkmark & && &0.1650 &0.5540 &3.4503 \\
2&\checkmark & & &\checkmark &\textbf{0.0413} &\underline{0.3249} &2.7587 \\
3&\checkmark & &\checkmark& &0.1197 &0.4264 &3.2072 \\
4&\checkmark & &\checkmark&\checkmark &0.0499 &0.3277 &2.5753 \\
5&\checkmark &\checkmark &&\checkmark &0.0502 &0.3432 &\underline{2.5154} \\
6&\checkmark &\checkmark && &0.1705 &0.4154 &2.7027 \\
7&\checkmark &\checkmark & \checkmark& &0.1450 &0.4087 &2.6509 \\
8&\checkmark &\checkmark &\checkmark&\checkmark &\underline{0.0419} &\textbf{0.3213} &\textbf{2.2943} \\
\hline
\end{tabular}
}
\label{tab:ablation}
\end{table}

\begin{table}[t]
\centering
    \caption{Impact of different spatial regularizations. Cluster-level regularization performs better than the KNN-based regularization. Bold fonts and underlines indicate the best and the second-best performance, respectively.}\label{tab:cluster}
    \renewcommand\arraystretch{1.1}

        \begin{tabular}{ll|ccc}
        \hline
        \multicolumn{2}{c|}{\multirow{2}*{Method}}&Static &Slow &Fast \\
        \cline{3-5}
        &&\multicolumn{3}{c}{Mean Errors $\downarrow$}\\
        \hline
        \multicolumn{2}{l|}{Baseline}&0.0502 &0.3432 &2.5154 \\
        \hline
        \multirow{3}*{+ $\mathcal{L}_{KNN}$} & $K=3$ &\textbf{0.0371} &0.3370 &2.5362 \\
        &$K=5$ &0.0490 &0.3378 &2.4568 \\
        &$K=10$ &0.0731 &0.3405 &2.5116 \\
        \hline
        \multirow{3}*{+ $\mathcal{L}_{c}$ (ours)}&$d_c = 2$ &0.0526 &0.3438 &\textbf{2.1814}
        \\
        &$d_c = 3$ &\underline{0.0419} &\textbf{0.3213} &\underline{2.2943} \\
        &$d_c = 4$ &0.0459 &\underline{0.3318} &2.4434 \\
        \hline
        \end{tabular}
    
\end{table}

\begin{table}[t]
\centering
    \caption{Varying $\theta_b$ for backward consistency loss. - indicates the results without the $\exp(-\frac{t}{\theta_b})$ term. Bold fonts and underlines indicate the best and the second-best performance, respectively.}\label{tab:thetab}
    \renewcommand\arraystretch{1.2}

        \begin{tabular}{l|ccc}
        \hline
        \multirow{2}*{$\theta_b$}&Static &Slow &Fast \\
        \cline{2-4}
        &\multicolumn{3}{c}{Mean Errors $\downarrow$}\\
        \hline
        $5$ &0.0482 &0.3264 &\underline{2.3580} \\
        $10$ &0.0419& \underline{0.3213}&\textbf{2.2943}\\
        $15$ &\textbf{0.0293}&\textbf{0.3091} &2.4149 \\
        $-$ &\underline{0.0382}& 0.3298& 2.4596\\
        \hline
        \end{tabular}
        
\end{table}

\begin{figure*}[t]
	\begin{center}
		\includegraphics[width=0.95\linewidth]{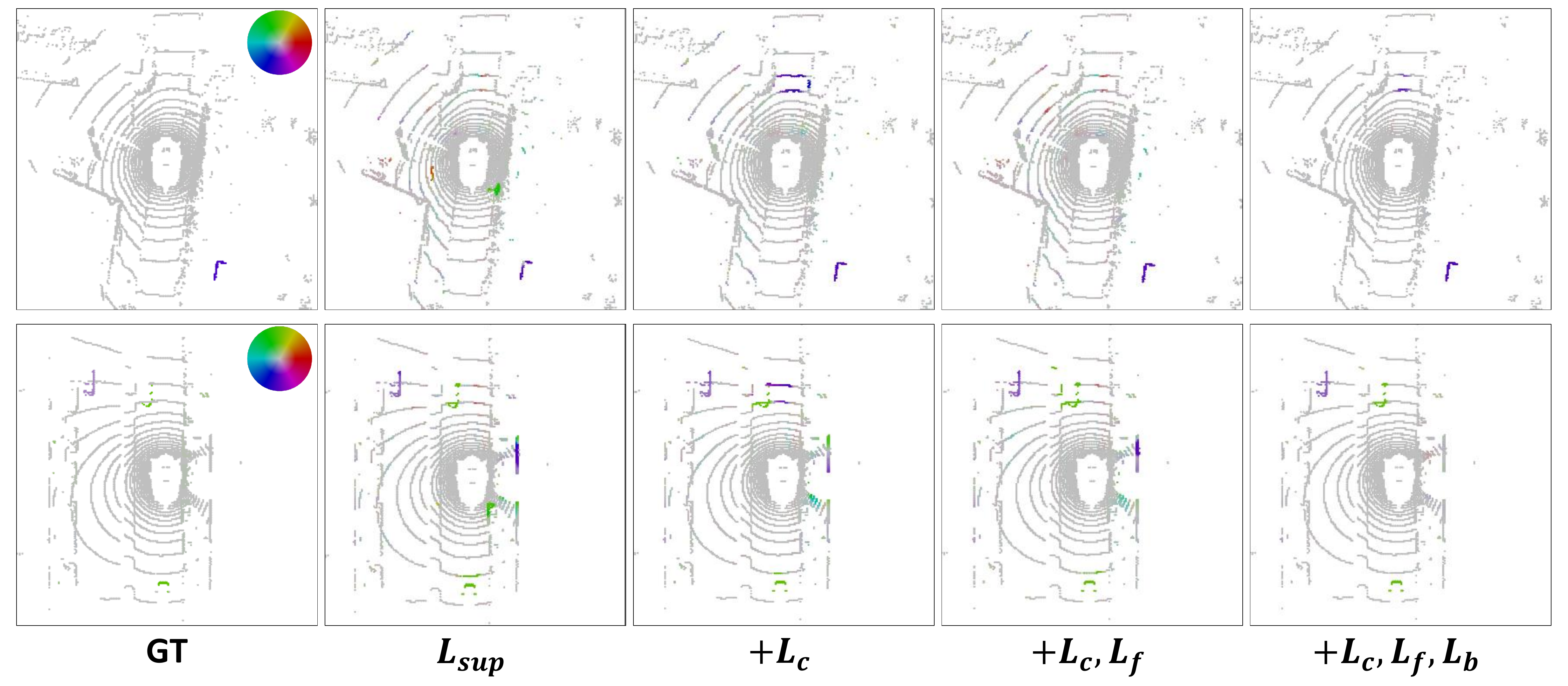}
		\caption{Qualitative results of the proposed self-supervised approach. The future displacements ($M^{T\rightarrow T+5}$) are depicted using the color wheel representation. (Zoom in for the best view)}
       \vspace{-15pt}
		\label{fig:qualitative}
	\end{center}
\end{figure*}

\noindent\textbf{Qualitative results.}
We present the result visualizations of our approach in  Fig.~\ref{fig:qualitative} to qualitatively show the impacts of each regularization term. When directly training with pseudo labels $\mathcal{L}_{sup}$, the model tends to produce inconsistent predictions for the same object and predict large motions for background cells. The third column illustrates that cluster consistency loss $\mathcal{L}_c$ effectively benefits spatial consistency. With the addition of forward regularization $\mathcal{L}_f$, the prediction results are further improved, but there are still noticeable noises in the background cells. Finally, by incorporating backward regularization $\mathcal{L}_{b}$, the noise predictions in the background cells are significantly suppressed.

\subsection{Results on Waymo Dataset}
Following WeakMotionNet~\cite{weakmotion}, we further evaluate our approach on the Waymo dataset~\cite{waymo}. As shown in Table~\ref{tab:waymo}, similar to the results on the nuScenes dataset, our approach achieves mean errors of around 0.05m, 0.4m, and 2.3m at static, slow, and fast speed levels, respectively, outperforming self-supervised WeakMotionNet.

\begin{table}[t]
    \centering
    \caption{Motion prediction results on the Waymo Dataset. Our method still outperforms self-supervised WeakMotion by a large margin.}
    \label{tab:waymo}
    \renewcommand\arraystretch{1.2}

    \begin{tabular}{l|c|ccc}
    \hline
    \multirow{2}*{\makecell[c]{Method}}&\multirow{2}*{\makecell[c]{Sup.}}&Static &Slow &Fast \\
    \cline{3-5}
    &&\multicolumn{3}{c}{Mean Errors $\downarrow$}\\
    \hline
    MotionNet~\cite{Wu2020MotionNetJP}&Full.& 0.0263 & 0.2620 &0.9493 \\
    WeakMotion~\cite{weakmotion} &Weak.& 0.0297& 0.3458 &1.5655 \\
    \hline
    WeakMotion~\cite{weakmotion}&Self.& 0.1345& 0.5833 &3.0180 \\
    Ours&Self.& \textbf{0.0515}& \textbf{0.4440} &\textbf{2.3692} \\
    \hline
    \end{tabular}
    
\end{table}

\subsection{Discussions}
\noindent \textbf{Ground Segmentation.} We evaluate the performance of the ground segmentation algorithm~\cite{lee2022patchwork++} on the nuScenes dataset, and get a 95\% precision rate while a 92\% recall rate, indicating the majority of ground points can be correctly removed. Although almost all ground points can be successfully removed, the self-supervised training process still suffers from remaining ground points, noises, and points without correspondences. The issues are alleviated using our proposed regularizations, as shown in Fig.~\ref{tab:ablation}.
\\
\noindent \textbf{Clustering.} In the paper, we utilize the breadth-first clustering (BFS) algorithm to demonstrate the effectiveness of the cluster consistency loss. We evaluate the clustering results of BEV cells on the nuScenes dataset and observe a 9.6\% under-clustering rate and an 11.1\% over-clustering rate. It indicates that although effective, the BFS algorithm still wrongly clusters cells from different instances in some cases. Therefore, fellow researchers can consider utilizing or developing more advanced clustering strategies, which may lead to better performances. 
\\
\noindent \textbf{Efficiency.} During training, ground segmentation and clustering take around 5ms and 10ms, respectively. In practice, these processes can be employed as data pre-processing before training. For the optimal transport (OT) solver, we implement it with Pytorch and operate in batch. It takes less than 1ms to solve the OT problem.
During inference, given that our approach is model-independent, the inference time is based on the model used. When applying MotionNet, it takes 10ms for point cloud transformation and voxelization and another 10ms for prediction~\cite{Wu2020MotionNetJP}.

\section{Conclusion}
In this paper, we introduce a novel approach for self-supervised motion prediction with only unlabeled LiDAR point clouds. We utilize optimal transport to generate pseudo labels and propose three novel spatial and temporal regularization terms to facilitate self-supervised motion learning. Extensive experiments on the large-scale nuScenes dataset illustrate that our approach significantly surpasses previous self-supervised methods and shallows the gap with fully-supervised ones.

\noindent\textbf{Acknowledgements.}
This study is supported under the RIE2020 Industry Alignment Fund – Industry Collaboration Projects (IAF-ICP) Funding Initiative, as well as
cash and in-kind contribution from the industry partner(s).
This research is also supported by the MoE AcRF Tier 2 grant (MOE-T2EP20220-0007) and the MoE AcRF Tier 1 grant (RG14/22).

{
    \small
    \bibliographystyle{ieeenat_fullname}
    \bibliography{refs}
}
\end{document}